%% file: main.tex
\documentclass[10pt,twocolumn,letterpaper]{article}

\usepackage[pagenumbers]{cvpr} 

\input{preamble}

\definecolor{cvprblue}{rgb}{0.21,0.49,0.74}
\usepackage[pagebackref,breaklinks,colorlinks,citecolor=cvprblue]{hyperref}

\usepackage{adjustbox}
\usepackage{utfsym}

\def\paperID{*****} 
\def\confName{CVPR}
\def\confYear{2024}

\title{Video Question Answering for People with Visual Impairments \\ Using an Egocentric 360-Degree Camera}

\author{
    Inpyo Song$^1$, Minjun Joo$^1$, Joonhyung Kwon$^{2}$, Jangwon Lee$^1$\\
    $^1$Department of Immersive Media Engineering, Sungkyunkwan University\\
    $^2$School of Electronics and Information Engineering, Korea Aerospace University\\
    \tt\small{\{songinpyo, jmjs1526, leejang\}@skku.edu, ludin9901@kau.kr}
}

\begin{document}
\maketitle

\begin{abstract}
This paper addresses the daily challenges encountered by visually impaired individuals,
such as limited access to information, navigation difficulties, and barriers to social interaction.
To alleviate these challenges, we introduce a novel visual question answering dataset.
Our dataset offers two significant advancements over previous datasets:
Firstly, it features videos captured using a 360-degree egocentric wearable camera,
enabling observation of the entire surroundings, departing from the static image-centric nature of prior datasets.
Secondly, unlike datasets centered on singular challenges,
ours addresses multiple real-life obstacles simultaneously through an innovative visual-question answering framework.
We validate our dataset using various state-of-the-art VideoQA methods and diverse metrics.
Results indicate that while progress has been made, satisfactory performance levels for AI-powered assistive services
remain elusive for visually impaired individuals.
Additionally, our evaluation highlights the distinctive features of the proposed dataset,
featuring ego-motion in videos captured via 360-degree cameras across varied scenarios.
\end{abstract}

\begin{figure}[!t]
  \centering
  \includegraphics[width=\columnwidth]{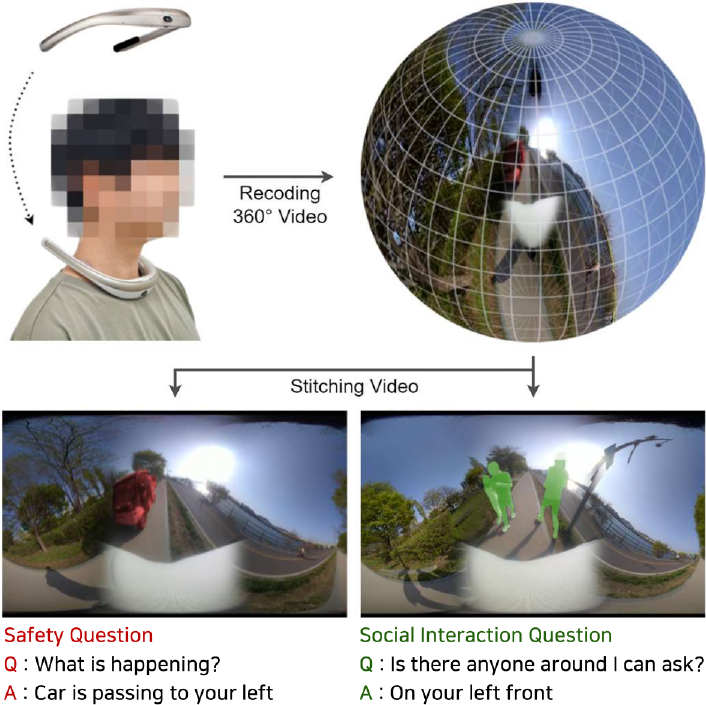}
  \caption{
We introduce a novel visual question answering dataset comprising videos captured with a wearable 360-degree camera,
aiming to address common challenges visually impaired individuals may encounter
by recording entire surroundings and providing VQA-style annotations for various situations. 
    }
   \label{fig1:front_fig}
\vspace{-1em}
\end{figure}

\section{Introduction} \label{Sec:Introduction}
Visually Impaired Persons (VIPs) face a variety of challenges in their daily lives
due to their limited access to visual information for understanding their surroundings.
Despite significant research efforts directed towards developing artificial intelligence-based assistive technologies
aimed at enhancing accessibility and independence for VIPs,
current solutions often encounter difficulties in adapting effectively to dynamic environments \cite{sivan2016computer}.
This difficulty arises from the fact that VIPs face a diverse range of obstacles,
while most current solutions, based on previously proposed datasets, focus on single specific tasks. 
These obstacles include identifying objects and text \cite{jafri2014computer},
navigating through environments \cite{bacsgoze2020navigational},
and interpreting nonverbal cues during social interactions \cite{romo2023examination}.
Additionally, safety remains a crucial concern,
as VIPs frequently encounter greater risks in everyday situations \cite{ahmed2016addressing}.
While the notable research project VizWiz has been introduced to develop algorithms for assistive technologies
aiding individuals who are blind, most tasks introduced with datasets in the project
rely on static images rather than videos \cite{gurari2018vizwiz}.
Moreover, some of these tasks require VIPs to manually capture photos for inquiries,
which can be impractical in real-world application scenarios with dynamic settings that demand rapid responses.

In response to these challenges, we introduce the
\textbf{VIEW-QA} dataset (\textbf{V}isually \textbf{I}mpaired \textbf{E}quipped
with \textbf{W}earable 360-degree camera \textbf{Q}uestion \textbf{A}nswering).
To construct this dataset, we collected videos using a 360-degree egocentric wearable camera
that can offer VIPs a continuous and hands-free visual feed, aiming to enhance its usability and interaction with their environment.
From these videos, we made a VideoQA dataset which contains annotations to comprehensively address varied needs.
Questions are categorized into \textit{Social Interaction, Vicinity Awareness, Object Information, Navigation,} and \textit{Safety}.
Each category is meticulously tailored to meet the specific challenges faced by VIPs,
making VIEW-QA an invaluable resource for
developing AI systems that can interpret complex visual scenes and deliver timely,
relevant information to assist VIPs in their daily lives.
Figure \ref{fig1:front_fig} shows the video collecting and annotation examples.
Furthermore, we have evaluated VIEW-QA using various established VideoQA methods,
including recent CNN-based models and Vision-Language Pretrained (VLP) models.


\section{Related Work} \label{Sec:Related works}

\subsection{Computer Vision for Visually Impaired People}
Recent advancements in computer vision have significantly enhanced support technologies for VIPs.
Research has mostly focused on a single specific challenge such as street crossings and indoor obstacle avoidance \cite{sivan2016computer,li2020cross,afif2020indoor}.
In the domain of VQA,
the VizWiz dataset incorporates images taken by VIPs,
addressing specific inquiries \cite{gurari2018vizwiz}.
Studies leveraging this dataset has led for ensuring privacy, analyzing reasoning,
and assessing necessary visual skills for query resolution \cite{bhattacharya2019does,zeng2020vision}.
However, these approaches often rely on static imagery and are limited to object-centric tasks,
revealing a gap in catering to the broader informational needs of VIPs through dynamic visual inputs.

\subsection{Video Question Answering}
VideoQA extends the scope from static images to dynamic video content,
presenting new challenges and opportunities \cite{Antol_2015_ICCV}.
MSVD-QA and MSRVTT-QA utilize video captioning for open-ended QA \cite{10.1145/3123266.3123427}, 
while TGIF-QA and ActivityNetQA emphasize spatio-temporal reasoning, 
often in simplistic settings \cite{jang2017tgif,yu2019activitynet}. 
NExT-QA underscores the need for causal and temporal reasoning in video content analysis \cite{xiao2021next}. 
NewsVideoQA applies textual analysis to news video QA \cite{jahagirdar2023watching}, 
and Pano-avqa integrates audio to explore spatial reasoning in 360-degree videos \cite{yun2021pano}. 
EgoTaskQA and QAEgo4D focus on egocentric video understanding focusing on task specific and efficient data storage for first-person perspectives \cite{jia2022egotaskqa,datta2022episodic}.
Despite some progress,
research has mainly focused on understanding video context through text rather than addressing real-world issues.
Unlike these datasets,
we aim to tackle multiple daily challenges faced by VIPs by integrating various tasks into our new dataset.

\begin{table}[!t]
\centering
\small
\begin{adjustbox}{max width=\linewidth}
\begin{tabular}{ c c c c c c }
\toprule
Dataset & Goal & Ego. & 360$^{\circ}$ & Videos (K) & Video Lengths (s) \\
\hline
MSVD-QA \cite{10.1145/3123266.3123427} & Descriptive QA & & & 2.0 & 10\\
MSRVTT-QA \cite{10.1145/3123266.3123427} & Descriptive QA & & & 10.0 & 15  \\
TGIF-QA \cite{jang2017tgif} & Spatio-temporal reasoning & & & 71.7 & 3.1 \\
ActivityNet-QA \cite{yu2019activitynet} & Descriptive QA & & & 5.8 & 180 \\
NExT-QA \cite{xiao2021next} & Causal \& temporal reasoning & & & 5.4 & 44 \\
NewsVideoQA \cite{jahagirdar2023watching} & Subtitle understanding & & & 3.0 & 10  \\
EgoTaskQA \cite{jia2022egotaskqa} & Task understanding & \usym{1F5F8} & & 2.0 & 25 \\
QAEgo4D \cite{datta2022episodic} & Episodic memory QA & \usym{1F5F8} & & 1.3 & 495.1 \\
Pano-AVQA \cite{yun2021pano} & Spatial reasoning & & \usym{1F5F8} & 5.4 & 5 \\
\textbf{VIEW-QA} (ours) & Supporting VIPs & \usym{1F5F8} & \usym{1F5F8} & 1.0 & 34.4 \\
\bottomrule
\end{tabular}
\end{adjustbox}
{
\caption{
Comparison of open-ended VideoQA datasets.
VIEW-QA stands out with its focus on egocentric 360-degree videos and objective to support VIPs by addressing practical challenges.
}
\label{tab:VQA dataset comparision}
}
\end{table}

\begin{figure}[!t]
  \centering
  \includegraphics[width=\columnwidth]{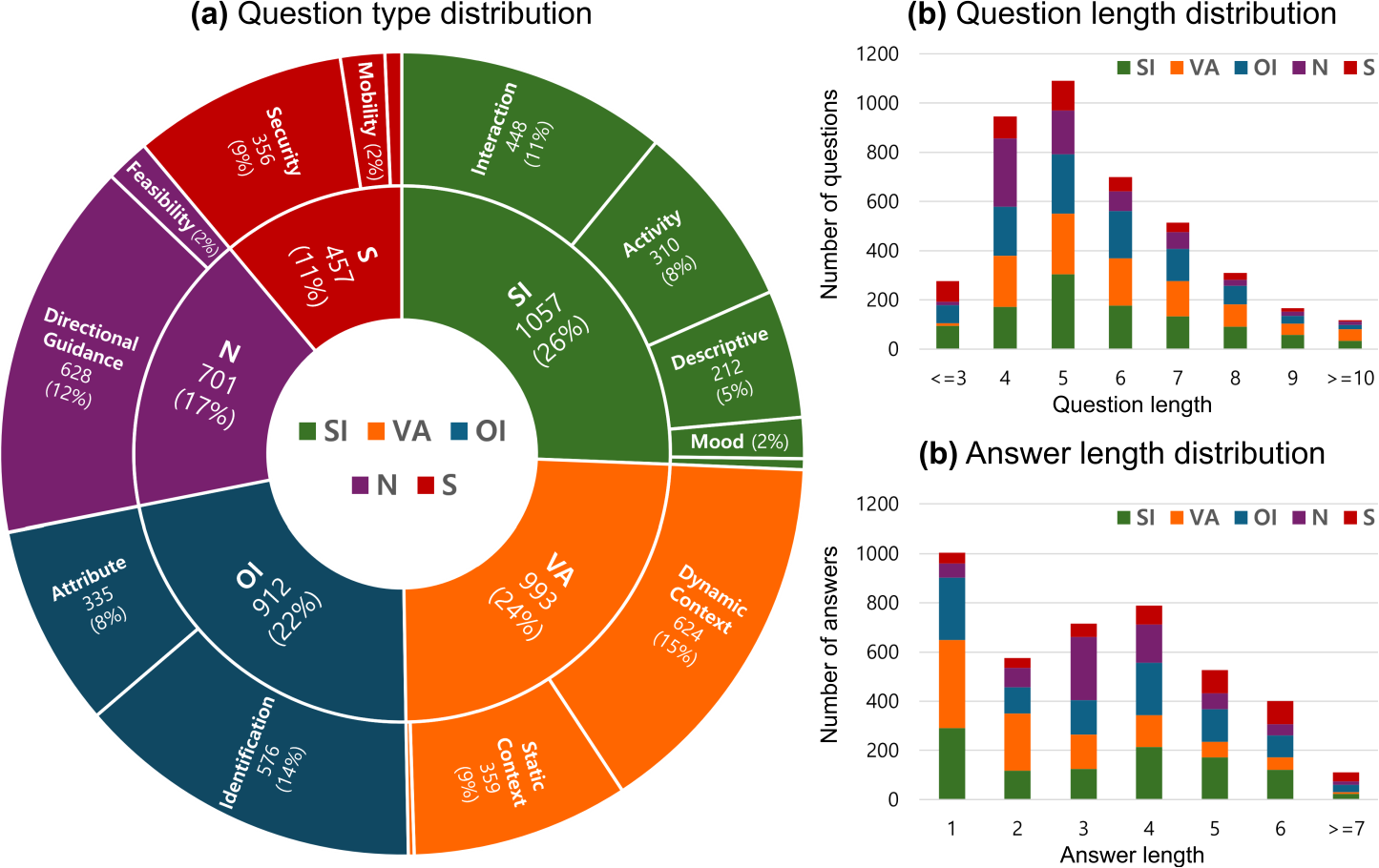}
  \caption{
  Overview of VIEW-QA dataset characteristics.
  (a) Distribution of the question types.
  (b) \& (c) Average question and answer lengths are 5.6 and 3.2 words, 
  designed for VIPs with concise questions and detailed answers from visual perception challenges.
    }
   \label{fig2:data_fig}
\vspace{-1em}
\end{figure}

\begin{table*}[!t]
\centering
\begin{adjustbox}{width=\linewidth}
\begin{tabular}{ l | c  c  c  c  c | c | c | c | c | c }
\toprule
Method & $\textit{WUPS}_{SI}$ & $\textit{WUPS}_{VA}$ & $\textit{WUPS}_{OI}$ & $\textit{WUPS}_{N}$ & $\textit{WUPS}_{S}$ & \textit{WUPS} & \textit{Cont.} & \textit{Acc.} & \textit{Llama2} & \textit{GPT-4} \\
\hline
Popular & 15.7 & 12.3 & 15.6 & 18.4 & 2.5 & 13.9 & 10.3 & 6.8 & 32.8 & 17.7 \\
BlindQA & 22.1 & 20.0 & 15.3 & 25.6 & 10.9 & 19.5 & 21.1 & 11.5 & 34.2 & 27.2 \\
HME \cite{fan2019heterogeneous} & 25.2 & 21.1 & 16.9 & 25.6 & 12.7 & 21.1 & 22.1 & 12.7 & 35.8 & 28.5 \\
CoMem \cite{gao2018motion} & 25.3 & 22.0 & 20.2 & 23.8 & 17.7 & 22.3 & 23.2 & 11.7 & 38.6 & 29.9 \\
UATT \cite{xue2017unifying} & 26.4 & 21.8 & 21.3 & 24.4 & 15.9 & 22.7 & 23.3 & 12.5 & 37.2 & 29.3 \\
EVQA \cite{Antol_2015_ICCV} & 24.3 & 23.6 & 21.6 & 24.9 & 18.6 & 23.0 & 23.5 & 12.8 & 37.3 & 31.8 \\
HGA \cite{jiang2020reasoning} & 24.0 & 25.8 & 24.5 & 16.5 & 21.4 & 23.2 & 23.5 & 12.3 & 35.5 & 27.9 \\
\hline
ViTis \cite{engin2023zero} & 25.9 & 23.4 & 28.2 & 26.3 & \textbf{21.5} & 25.5 & 27.7 & \textbf{14.6} & \textbf{45.2} & \textbf{35.1} \\
EgoVLPv2 \cite{pramanick2023egovlpv2} & \textbf{31.0} & \textbf{27.2} & \textbf{31.0} & \textbf{27.2} & 16.7 & \textbf{28.0} & \textbf{27.9} & 13.0 & 43.1 & 34.1 \\
\bottomrule
\end{tabular}
\end{adjustbox}
{
\caption{
Comparative analysis of VideoQA methods on VIEW-QA dataset.
}
\label{tab:VIEWQA experiment}
}
\end{table*}

\section{VIEW-QA Dataset} \label{Sec:Dataset}
Existing datasets typically address narrow aspects of challenges, 
failing to encompass the broad range of real-life concerns that VIPs face. 
The VIEW-QA dataset addresses this by providing a comprehensive collection 
that uses 360-degree egocentric videos to extensively cover essential VIP concerns, 
supported by specifically tailored questions.
\subsection{Question Types}
The questions are segmented into five critical areas identified as primary challenges for VIPs.
\begin{itemize}
    \item \textbf{Social Interaction (SI).} 
    Questions aimed at helping VIPs recognize emotional states, social activities, and interactions, and descriptions of individuals, facilitating more effective communication and social engagement.
    \item \textbf{Vicinity Awareness (VA).}
    Exploratory questions about incidental environmental features which can assist VIP to get better comprehensive awareness in their vicinity.
    \item \textbf{Object Information (OI).} 
    Focuses on aiding VIPs in recognizing everyday objects through attributes like color, shape, and labels, thus improving their ability to interact with their surroundings independently.
    \item \textbf{Navigation (N).} 
    Provides information critical for safely and efficiently navigating spaces, 
    such as directions and the identification of obstacles.
    \item \textbf{Safety (S).} Addresses safety concerns by identifying potential hazards in the VIP's environment, ranging from immediate physical dangers to more subtle security risks.
\end{itemize}
The distribution of these question types within the dataset is illustrated in Figure \ref{fig2:data_fig}.

\subsection{Video Collection}
To ensure the VIEW-QA dataset mirrors the real-world experiences of VIPs,
we conducted a comprehensive literature review on their primary concerns \cite{ahmed2016addressing,jafri2014computer, bacsgoze2020navigational, romo2023examination}
and collaborated with institutions specializing in VIP services.
We captured footage in typical settings encountered by VIPs, 
including ATMs, subways, streets, cafes, and markets, 
using a 360-degree wearable camera worn by an actor simulating VIP experiences.
The dataset comprises 1,030 videos containing approximately 10 hours and 1,062,960 frames.

\subsection{Annotations}
The annotations of questions and answers were performed
by separate annotators trained to comprehend the specific concerns and needs of VIPs.
This training included a thorough review of the relevant literature 
to ensure a deep understanding of the context.
To ensure the quality and relevance of the annotations, the following guidelines were rigorously applied:
\textbf{Avoid visual clues.} Questions must not contain words that could serve as visual clues.
\textbf{Relevance to VIPs.} QAs should specifically target information that is essential and relevant to VIPs.
\textbf{Intention awareness.} Answers must be annotated with an awareness of the question's intent, as defined by the question types.
\textbf{Exclusion criteria.} Any video or question deemed inappropriate or not informative should be excluded from the dataset.
We organized 4 QAs per video, resulting in a total of 4,120 annotated QA pairs.

\subsection{Privacy and Ethics}
The collection of the VIEW-QA dataset was conducted under Institutional Review Board (IRB) approval, adhering to strict ethical standards.
Informed consent was secured from all participants, and privacy measures like blurring and facial masking were implemented for non-consenting individuals in the videos.
Careful measures were also made to avoid capturing sensitive or personal activities.

\section{Experiments} \label{Sec:Experiments}

\subsection{Experimental Setup}
\noindent \textbf{Baselines.} 
To validate the effectiveness of the VIEW-QA dataset,
we benchmarked against established CNN-based models \cite{Antol_2015_ICCV,gao2018motion,fan2019heterogeneous,xue2017unifying,jiang2020reasoning} 
and VLP models \cite{engin2023zero,pramanick2023egovlpv2}. 
We also included simpler baselines: 
Popular, which selects the most frequent answer per question type, 
and BlindQA, using only textual information without visual input.

\noindent \textbf{Evaluation metrics.} 
Accuracy (\textit{Acc.}), commonly used in VQA and VideoQA, 
but sometimes overlooks semantically correct but textually varied responses \cite{ging2024open}.
To enhance evaluation precision,
we used two more flexible metrics and employed language models for judgment.
The Wu-Palmer Similarity (\textit{WUPS}) assesses word similarity through WordNet \cite{malinowski2014multi}, 
and the Contains (\textit{Cont.}) metric verifies if essential content of the ground truth is present in the predicted answer, 
excluding stop words.
We also employed large language models, 
\textit{Llama2} and \textit{GPT-4}, 
to score semantic similarity between predictions and ground truth, 
enhancing alignment with human judgment \cite{touvron2023llama,achiam2023gpt}.

\noindent \textbf{Experimental results.}
Despite the superior performance of recent AI models compared to our two simple baselines (Popular and BlindQA),
as indicated in Table \ref{tab:VIEWQA experiment},
their overall performance remains below satisfactory levels across all evaluation metrics.
Among the two different types of models, VLP models outperformed CNN-based models.
Specifically, EgoVLPv2 excelled $\textit{WUPS}$ and $\textit{Cont.}$ metrics,
effectively capturing essential keywords.
On the other hand, ViTis achieved the best results in $\textit{Acc.}$, $\textit{Llama2}$, and $\textit{GPT-4}$, 
demonstrating its ability to generate answers that are either exactly correct or semantically close with the ground truth.

\section{Conclusion} \label{Sec:Conclusion}
We introduce \textbf{VIEW-QA}, the first egocentric 360-degree VideoQA dataset for visually impaired individuals (VIPs).
This dataset addresses the various daily challenges they often face with questions deliberately crafted to mirror their daily experiences.
VIEW-QA establishes a benchmark in assistive technology,
designed to augment VIP autonomy through dynamic visual interpretation.
Future research will expand the dataset’s scope and refine methods
to effectively identify intended clues without visual cues in the questions.
\newline
\noindent\textbf{Acknowledgement:} This work was supported by the National Research Foundation of Korea (NRF)
grant funded by the Korea government(MSIT) (No. 2021R1G1A1010720).

{
    \small
    \bibliographystyle{ieeenat_fullname}
    \bibliography{main}
}

\end{document}

%% file: preamble.tex
\usepackage[dvipsnames]{xcolor}
